\documentclass[runningheads]{llncs}

\usepackage{eccv}

\usepackage{eccvabbrv}

\usepackage{graphicx}
\usepackage{booktabs}

\usepackage[accsupp]{axessibility}  
\usepackage{adjustbox}
\usepackage{booktabs,ragged2e,amsmath,tabularx}
\newcolumntype{C}{>{\Centering\hspace{0pt}}X}
\usepackage{multirow}
\usepackage{booktabs}
\usepackage{array}
\usepackage{caption}
\usepackage{colortbl}  
\usepackage{xcolor}    
\usepackage{graphicx}
\usepackage{subcaption}
\usepackage{enumitem}
\usepackage{amsmath}
\usepackage{amssymb}

\usepackage[pagebackref,breaklinks,colorlinks,citecolor=eccvblue]{hyperref}

\usepackage{orcidlink}

\begin{document}

\title{EUFCC-CIR: a Composed Image Retrieval Dataset for GLAM Collections} 

\titlerunning{EUFCC-CIR: a Composed Image Retrieval Dataset for GLAM Collections}

\author{Francesc Net\orcidlink{0000-0001-6888-519X} \and
Lluis Gomez\orcidlink{0000-0003-1408-9803}}

\authorrunning{F. Net et al.}

\institute{Computer Vision Center, Universitat Autònoma de Barcelona.\\
\email{\{fnet,lgomez\}@cvc.uab.cat}}

\maketitle

\begin{abstract}
  The intersection of Artificial Intelligence and Digital Humanities enables researchers to explore cultural heritage collections with greater depth and scale. In this paper, we present EUFCC-CIR, a dataset designed for Composed Image Retrieval (CIR) within Galleries, Libraries, Archives, and Museums (GLAM) collections. Our dataset is built on top of the EUFCC-340K image labeling dataset and contains over 180K annotated CIR triplets. Each triplet is composed of a multi-modal query (an input image plus a short text describing the desired attribute manipulations) and a set of relevant target images. The EUFCC-CIR dataset fills an existing gap in CIR-specific resources for Digital Humanities. We demonstrate the value of the EUFCC-CIR dataset by highlighting its unique qualities in comparison to other existing CIR datasets and evaluating the performance of several zero-shot CIR baselines. The dataset is publicly available at \url{https://github.com/cesc47/EUFCC-CIR}. 
\end{abstract}

\section{Introduction}
\label{sec:intro}


The integration of artificial intelligence methodologies with digital humanities has introduced new opportunities for data search, analysis, and visualization, significantly enhancing our capabilities to understand large collections of cultural heritage assets. In this context, image retrieval is a key task, allowing users to quickly search and locate visual content in large collections. Image retrieval can be approached in several ways, depending on the nature of the queries that are used to express the user's intent. For example, we refer to content-based image retrieval when the query information is an image and we expect to find similar images, while text-to-image retrieval involves queries expressed in natural language. Cross-modal retrieval enables the use of different types of queries, such as images and text, to find related content across modalities.

\begin{figure}[t]
    \centering
    \includegraphics[width=\textwidth]{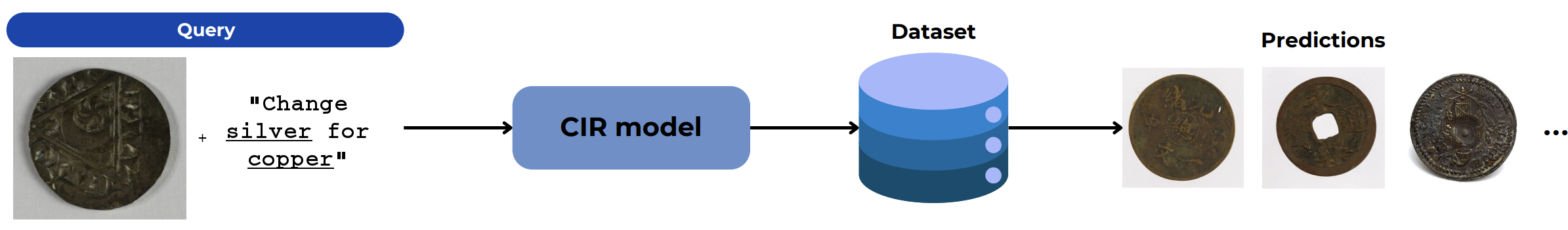}
    \caption{Composed Image Retrieval (CIR) example. The user query is expressed with two modalities: an image of a silver coin and a short text (``Change silver for copper'') that describes the desired modifications. These inputs are processed by the CIR model, which searches in a dataset to generate a ranking of predictions according to the visual and textual modalities.}
    \label{fig:overview_cir}
\end{figure}

One emerging application within this landscape is Composed Image Retrieval (CIR)~\cite{vo2019composing,han2022uigr,baldrati2022conditioned,neculai2022probabilistic,saito2023pic2word}, where the query consists of an image along with a short descriptive text that describes desired modifications to the input image (see example in Figure \ref{fig:overview_cir}). This task is particularly relevant for Galleries, Libraries, Archives, and Museums (GLAM) collections, where users often seek specific variations or enhancements of known cultural artifacts. Despite the potential of CIR in enriching user experience and facilitating scholarly research in the GLAM domain, there remains a scarcity of datasets tailored for this purpose.

In this paper, we introduce EUFCC-CIR, a novel dataset designed explicitly for the task of Composed Image Retrieval within GLAM collections. Derived from the EUFCC-340K dataset~\cite{net2024eufcc340kfacetedhierarchicaldataset}, EUFCC-CIR leverages automated processes and filtering heuristics to create a rich and diverse set of query-response pairs. The original EUFCC-340K dataset, compiled using the Europeana portal's REST API, encompasses over 346,000 annotated images from a wide array of European cultural artifacts. Each image is labeled with metadata aligned to the Getty Art \& Architecture Thesaurus (AAT), offering a robust foundation for AI applications in the GLAM domain.

Our approach to constructing the EUFCC-CIR dataset involves identifying images of the same ``object type'' with variations in their ``materials'' annotations, and vice versa. This method ensures that the dataset captures meaningful differences that are essential for CIR tasks. By applying heuristics and filters, we enhance the dataset's richness, ensuring that each element tuple is unique and varied. The dataset is divided into training, validation, and two test splits, enabling comprehensive evaluation and comparison of CIR models.

We demonstrate the usefulness of the EUFCC-CIR dataset through qualitative and quantitative analyses, highlighting its value for Digital Humanities research. Our results underscore the potential of CIR to disrupt how we interact with and interpret cultural heritage collections, providing deeper insights and enhancing accessibility.


\section{Related Work}
\label{sec:sota}
In recent years there has been a growing interest in new models for Composed Image Retrieval (CIR), where the input query is specified in the form of an image plus some text that describes desired modifications. 
In this section we present an overview of relevant CIR methods and related datasets.

Vo \etal~\cite{vo2019composing} proposed a model that makes use of a gate mechanism to bring closer the embedding spaces of the image plus text query and the target image. The proposed method is named  TIRG, an abbreviation standing for ``Text Image Residual Gating''. Building on the idea of specialized retrieval systems, Han \etal~\cite{han2022uigr} proposed a Unified Interactive Garment Retrieval (UIGR) method that exhibits a specialization in the domain of garment retrieval, predicated upon the consideration of two distinct contributing factors: Text-guided garment retrieval (TGR) and Visually compatible garment retrieval (VCR). 

Baldrati \etal~\cite{baldrati2022conditioned} proposed an approach that encompasses a dual-stage process. In the first stage, they fine-tuned the CLIP (Contrastive Language Image Pretraining)~\cite{radford2021learning} text encoder to adapt its embedding space to the CIR task. In the second stage, they trained a Combiner network which learns to fuse the multi-modal features extracted with the CLIP image encoder and the fine-tuned CLIP text encoder. The Combiner's purpose is to be able to move between two points in the image space using the textual information as displacement vectors in the image embedding space. Neculai \etal~\cite{neculai2022probabilistic} investigated composing multiple multi-modal queries in image retrieval. Rather than fusing embeddings from different modalities through a fixed set of learnable parameters, they parameterize each input as a probabilistic embedding~\cite{chun2021probabilistic} following a multivariate Gaussian, and composed embeddings by deriving a composite multivariate Gaussian based on a probabilistic composition rule. 

Up to here, all the reviewed methods are supervised models and hence require specially annotated training data for the Composed Image Retrieval (CIR) task. Pic2Word~\cite{saito2023pic2word} is the first zero-shot model proposed for CIR. It utilizes a pre-trained CLIP model which aligns images with their corresponding captions while distinguishing unpaired instances. The output embeddings of the two encoders (image and text) are aligned concerning each other's modality. In Pic2Word a mapping network is learned to project images to pseudo-language tokens so that CIR multimodal queries can be expressed in a single modality (text) to simplify the retrieval process.

Baldrati \etal \cite{baldrati2023zero} presented SEARLE, a zero-shot method to map reference image features to pseudo-word tokens in the CLIP embedding space and then combine them with captions for effective image retrieval. This innovative approach thus totally obviates the need for labeled training data and can be adapted to quite several domains. This view is complemented by Karthik \etal \cite{karthik2023vision}, who employed pre-trained vision-language models and large language models to be able to both generate and modify captions for query images without specialized training. This therefore extends the range of applicability for compositional image retrieval systems. On the other hand, ComposeAE~\cite{anwaar2021compositional} uses an autoencoder-based approach to learn text and image query composition, enforcing deep metric learning with additional rotational symmetry constraints to enhance retrieval accuracy.

Elaborating in a more detailed approach, Hosseinzadeh \etal \cite{hosseinzadeh2020composed} represents images as sets of local areas, establishing explicit relationships between words in modification texts and image regions, with great enhancements in retrieval performance.

A Dual Composition Network was proposed by Kim \etal \cite{kim2021dual}, it combines the Composition Network, which merges image and text features with the Correction Network, in that it models the difference between reference and target images. Dual approaches enhance robustness and preciseness in retrieval processes. On the other hand, Tial \etal \cite{tian2023fashion} presented Additive Attention Compositional Learning (AACL), a multi-modal transformer architecture designed to model image-text contexts effectively and has achieved state-of-the-art results in fashion image retrieval. 

With the compositional reasoning now indispensable in vision-language models, Ray \etal \cite{ray2024cola} underline a benchmark that demands a model to retrieve an image based on complex configurations of objects and attributes. It specifically emphasizes the challenges posed by compositional tasks and underlines the requirement that models must understand fine-grained relations between text and image components.

\subsection{CIR Datasets}
One of the main challenges in the field of CIR is the requirement for  large-scale annotated data, which is typically costly and laborious to collect. In table~\ref{tab:datasets} we present an overview comparison of several representative datasets in the CIR field with our EUFCC-CIR collection, which aims at the digital humanities domain.

\begin{table}[h!]
\centering
\begin{adjustbox}{width=\textwidth}
\begin{tabular}{>{\centering\arraybackslash}m{3.5cm} >{\centering\arraybackslash}m{6cm} >{\centering\arraybackslash}m{2cm} >{\centering\arraybackslash}m{2cm}}
\toprule
\textbf{Dataset} & \textbf{Characteristics} & \textbf{\# Images} & \textbf{\# Triplets} \\ \midrule
FashionIQ\cite{wu2021fashion} & Fashion image dataset with human annotations for compatibility and diversity. & 62K & 24K \\ \midrule
MIT-States\cite{isola2015discovering} & Dataset of images with states and transformations, e.g., apple-sliced, etc. & 54K & 68K \\ \midrule
Fashion200k\cite{han2017automatic} & Fashion dataset with 200k images and corresponding textual descriptions. & 200K & 200K \\ \midrule
Shopping100k\cite{ak2018efficient} & Fashion dataset of individual clothing items extracted from different e-commerce providers. & 101K & 16K \\ \midrule
CIRCO\cite{baldrati2023zero} & Composed image retrieval on Common Objects in context & 122K & 1K \\ \midrule
CIRR\cite{liu2021image} & Pairs of crowd-sourced, open-domain images with human-generated modifying text. & 21K & 36K \\ \midrule
CSS\cite{vo2019composing} & Synthetic dataset of images containing several different geometric objects (sphere, cube, etc.) sitting in a variety of layouts. & 37K & 32K \\ \arrayrulecolor{gray}\midrule
\rowcolor{lightgray} EUFCC-CIR (Ours) & Composed image retrieval based on EUFCC-340K dataset. & 340K & 180K \\ \arrayrulecolor{black}\bottomrule
\end{tabular}
\end{adjustbox}
\vspace{1em}
\caption{Overview of the datasets considered in this study, including their characteristics, the number of images, and the number of triplets, including the EUFCC-CIR dataset, which is introduced in this work.}
\label{tab:datasets}
\end{table}

\begin{figure}
    \centering
    \begin{tabular}{>{\centering\arraybackslash}m{2.75cm} >{\centering\arraybackslash}m{1.75cm} >{\centering\arraybackslash}m{5cm} >{\centering\arraybackslash}m{1.75cm}}
    \toprule
         FashionIQ\cite{wu2021fashion} & \includegraphics[width=\linewidth]{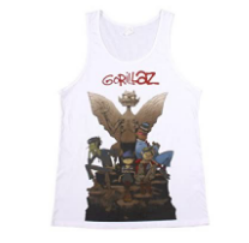} & + \texttt{cream with blue picture} = & \includegraphics[width=\linewidth]{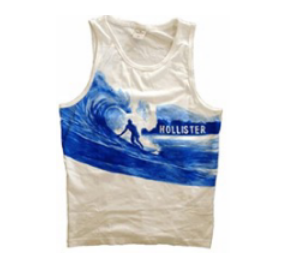} \\
         \midrule
         MIT-States\cite{isola2015discovering} & \includegraphics[width=\linewidth]{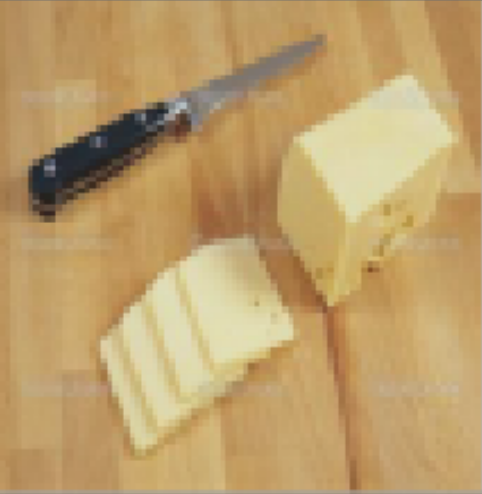} & + \texttt{change state to melted} = & \includegraphics[width=\linewidth]{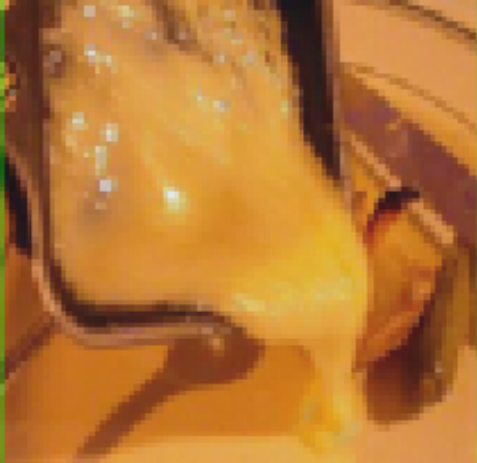} \\
         \midrule
         Shopping100k\cite{ak2018efficient} & \includegraphics[width=\linewidth]{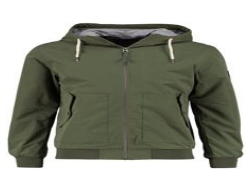} & + \texttt{with mandarin collar} = & \includegraphics[width=\linewidth]{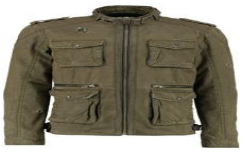} \\
         \midrule
         CIRCO\cite{baldrati2022conditioned} & \includegraphics[width=\linewidth]{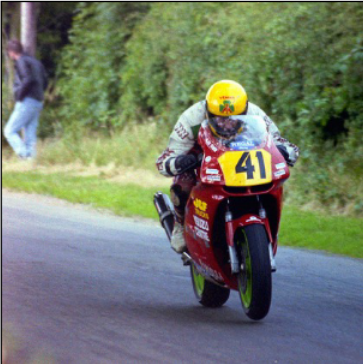} & + \texttt{is on track and has the front wheel in the air} = & \includegraphics[width=\linewidth]{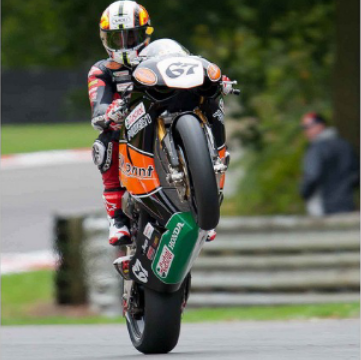} \\
         \midrule
         CIRR\cite{liu2021image} & \includegraphics[width=\linewidth]{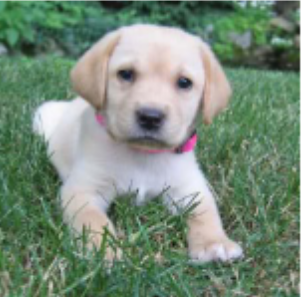} & + \texttt{the dog is inside a hole and not on grass} = & \includegraphics[width=\linewidth]{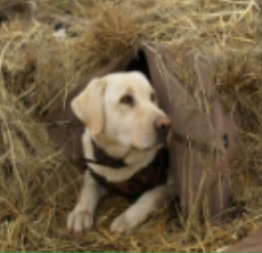} \\
         \midrule
         CSS\cite{vo2019composing} & \includegraphics[width=\linewidth]{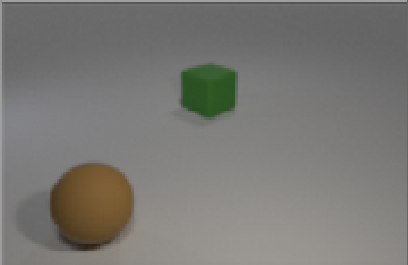} & + \texttt{add red sphere to top-left}~= & \includegraphics[width=\linewidth]{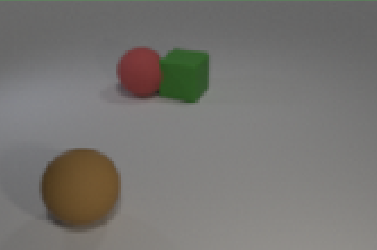} \\
         \midrule
         \rowcolor{lightgray} EUFCC-CIR (Ours) & \includegraphics[width=\linewidth]{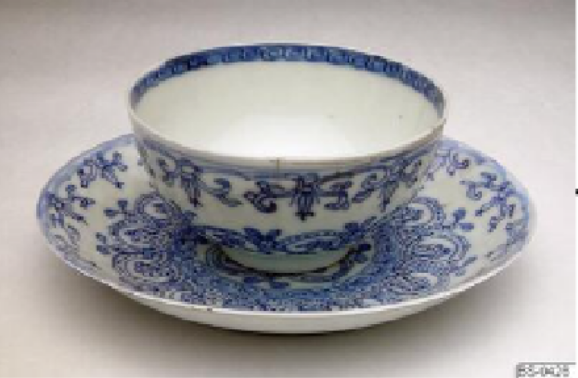} & + \texttt{change ceramic for porcelain}~= & \includegraphics[width=\linewidth]{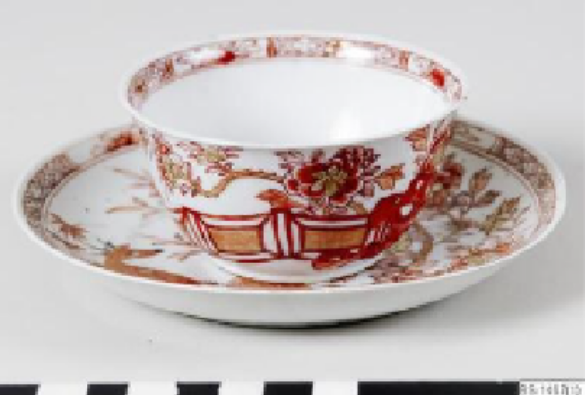} \\
    \bottomrule
    \end{tabular}
    \caption{Sample triplets from different CIR datasets.}
    \label{fig:db_samples}
\end{figure}

Figure~\ref{fig:db_samples} shows sample triplets from different CIR datasets, highlighting the differences in both visual domains and textual modifiers scope. We appreciate how datasets like FashionIQ\cite{wu2021fashion}, MIT-States\cite{isola2015discovering}, and Shopping100k\cite{ak2018efficient} focus on attribute manipulation, while CIRCO\cite{baldrati2022conditioned}, CIRR\cite{liu2021image}, and CSS\cite{vo2019composing} aim at generic scene manipulation by text description.

EUFCC-CIR is the first dataset for CIR in GLAM collections. 
Our dataset is built on top of the EUFCC-340K~\cite{net2024eufcc340kfacetedhierarchicaldataset}, an image multi-label categorization dataset collected from the Europeana portal\footnote{\url{http://europeana.eu}}, which is an online aggregator of collections of cultural heritage from across Europe. The collection comprises annotated images arising from cultural artifacts, multimedia content, and traditional records from a variety of European institutions. All the metadata of each item in EUFCC-340K contains rich detail for analytics, according to a hierarchical labeling structure following the Getty ``Art \& Architecture Thesaurus'' (AAT)\footnote{\url{https://www.getty.edu/research/tools/vocabularies/aat/}}. 

The EUFCC-340K initial data collection involved keyword searches and filtering for broad categories. Results were refined to include only those entries with available thumbnails and tagged with Reusability: OPEN, so the dataset comprises only images suitable for open research and application. The mapping of Europeana concepts to Getty AAT enabled structured labeling under four facets: ``Materials'', ``Object Types'', ``Disciplines'', and ``Subjects''. Manual curation ensured the quality of the dataset, although there might be remaining noisy annotations. Information on the data provider is given in each record. 

In total, the EUFCC-340K dataset has 346,324 annotated images from 358 data providers. Some data providers have massive stock whereas others have hardly contributed anything. The annotations for this dataset are partial and of varying detail. The dataset provides two different test subsets: The "Inner" test set has similar visual contents as the training and validation sets, while the "Outer" set is designed to stress models' performance with less represented visual content. This was done using a split strategy aimed at balance and diversifying according to the tag frequencies and also by setting appropriate minimum thresholds in each category.

\section{The EUFCC-CIR Dataset}
\label{sec:dataset}
In this work, we present the EUFCC-CIR dataset, where we repurpose the EUFCC-340K~\cite{net2024eufcc340kfacetedhierarchicaldataset} database to build CIR triplets composed of query images, text modifiers, and target images. While manually annotated CIR datasets usually consider that a query only has one potential target, our approach considers the fact that a query could have multiple possible targets because we use an automatic process to create the queries' image-text pairs and target images. The domain-specific challenges of the EUFCC-340K dataset render the task of CIR challenging and therefore, the proposed dataset aims to improve the robustness and practical applicability of CIR models in real-world scenarios.

The process of CIR triplet generation consists of an iterative loop, in which possible queries and target images are evaluated and selected according to a well defined criteria. 
Each image in the EUFCC-340K dataset has multi-label and hierarchical annotations across four different facets: ``Materials'', ``Object Types'', ``Disciplines'', and ``Subjects''. For our CIR dataset we focused on the ``Materials'' and ``Object Type'' attributes because they have richer hierarchical structures and vast vocabularies. Other attributes such as ``Subjects'' and ``Disciplines'' were ignored because they lacked hierarchical depth and variability.

\begin{figure}[h!]
    \centering
    \begin{subfigure}[b]{0.55\textwidth}
        \centering
        \includegraphics[width=\textwidth]{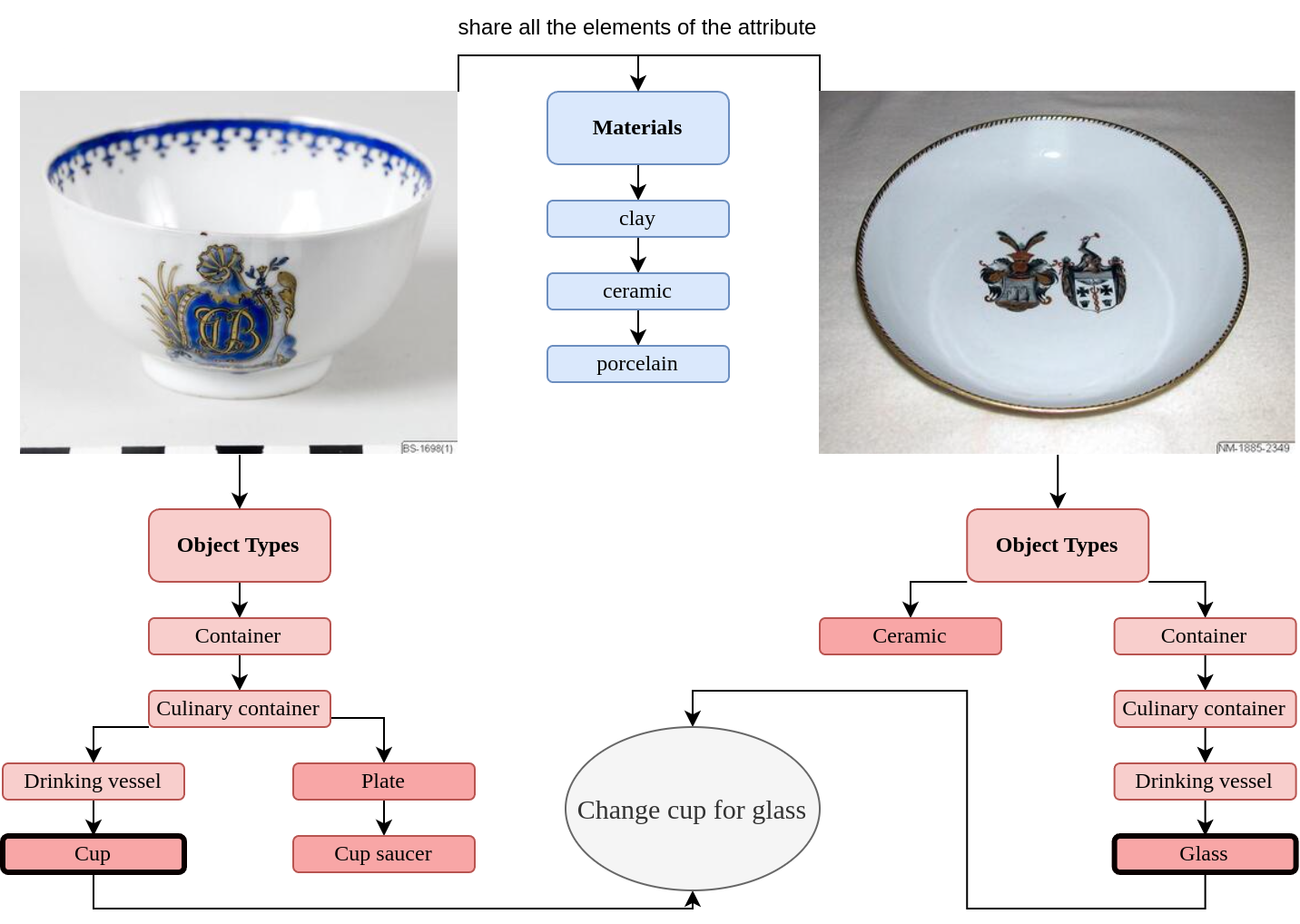}
        \caption{An example of a triplet found in the object types attribute hierarchy.}
        \label{fig:pairs_selection_sub1}
    \end{subfigure}
    \vfill
    \begin{subfigure}[b]{\textwidth}
        \centering
        \includegraphics[width=\textwidth]{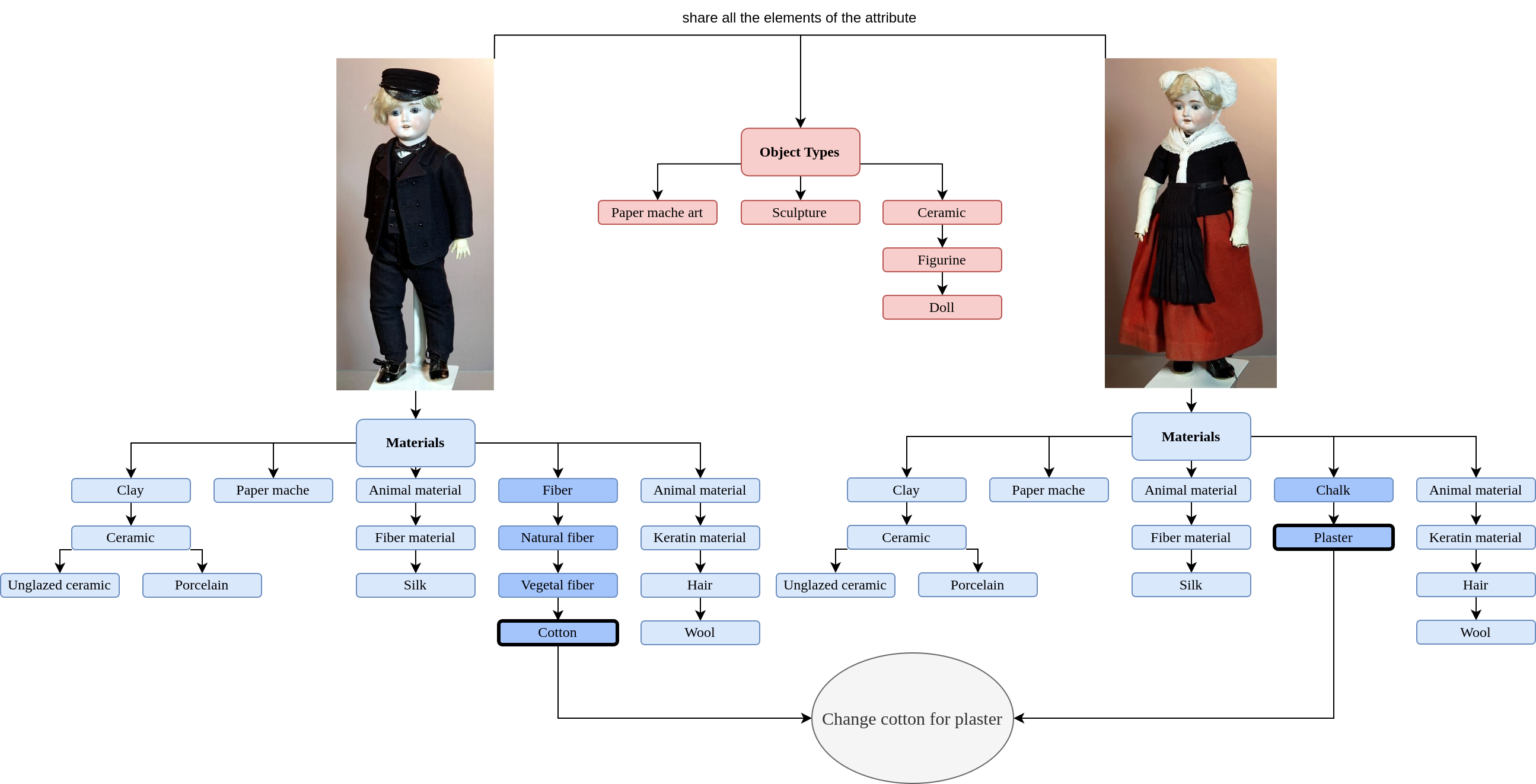}
        \caption{An example of a triplet found in the materials attribute hierarchy.}
        \label{fig:pairs_selection_sub2}
    \end{subfigure}
    \caption{Illustrative examples of how the EUFCC-CIR dataset triplets are ccreated, by analyzing the elements of the annotation hierarchy of each image attribute.}
    \label{fig:pairs_selection}
\end{figure}

As illustrated in Figure~\ref{fig:pairs_selection}, the underlying idea for creating CIR triplets from the EUFCC-340K dataset is to find image pairs that share significant similarities across their annotations while also exhibiting slight variations. To achieve this, we first select images in which the ``Materials'' or ``Object Type'' attributes are identical. Then, we iterate through their annotation hierarchical trees. If the attributes differ but have some commonalities, we explore the different leaves of the annotation trees, selecting different attributes from each other. Within the tree, siblings or descendants are chosen to maintain a reasonable semantic meaning for the relationship/modification. Additionally, several filters are applied to avoid redundancy in the dataset and ensure attribute diversity.

In Figure \ref{fig:pairs_selection_sub1} two images with a shared hierarchical annotation in the ``Materials'' attribute, specifically porcelain, are displayed. Despite this shared attribute, the images differ in their ``Object Type'' attribute. The differences are highlighted with darker red background nodes, while the elements selected to create one triplet are highlighted with a black border node.

To generate a query, we extract the differing elements from the query image and the target image and construct a sentence following the pattern \texttt{``change <\textit{query\_element}> for <\textit{target\_element}>''}. For instance, in this example, we select two descendants of the ``drinking vessel'' category (``glass'' and ``cup''), which are siblings nodes in the annotation hierarchy. Notice that for the same pair of images other triplets might be created, involving changing ``cup'' to another descendant of the ``culinary container'' category, such as a ``plate'' or ``cup saucer'' to form the triplet. The example in the figure illustrates only the generation of a single triplet.

In Figure \ref{fig:pairs_selection_sub2} the structure is examined from the perspective of the ``Object Type'' attribute. Here, the images share the same object type but differ in the ``Materials'' attribute. The unique distinct leaf in the materials hierarchy is highlighted with black bordered nodes, indicating the distinct attribute used to form the triplet. By selecting this different leaf, a meaningful triplet is generated, reflecting a subtle yet significant variation between the query and target images.

In order to ensure that the dataset is diverse and representative, attribute labels are constrained in terms of their iteration so that we do not clutter the data with too much of the same information. The re-use of the same image is also controlled to not over-represent the dataset. These constraints work to satisfy the size of the training set, which has a high volume of samples. However, this value can be changed such that fewer or more samples are generated in either case. This hierarchical structure, together with controlled query-target image pair selection, help in creating a high-quality CIR dataset.

During the EUFCC-CIR triplet generation, we follow the same dataset split strategy as in the official EUFCC-340K partition. We maintain the subset of image query-target pairs within members of a specific set. To clarify, triplets from the inner test set are generated using only items from the same set. This strict adherence to subset boundaries guarantees that models are trained, validated, and tested on different data partitions, encouraging generalization and avoiding data leakage.

\begin{figure}[h!]
    \centering
    \begin{subfigure}[b]{0.36\textwidth}
        \centering
        \includegraphics[width=\textwidth]{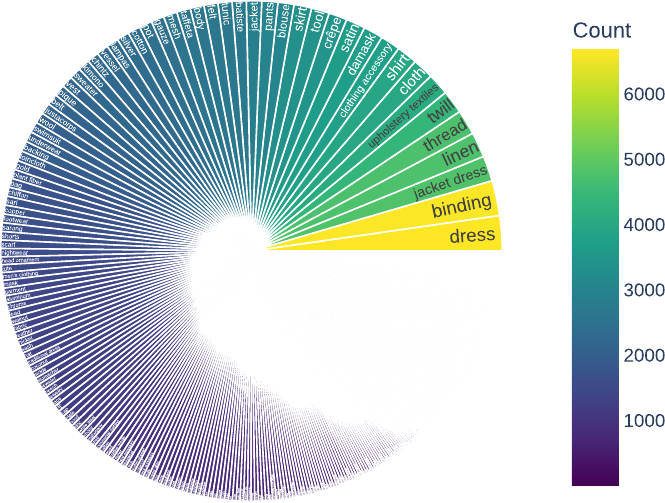}
        \caption{Training set.}
        \label{fig:train_sunburst}
    \end{subfigure}
    \hfill
    \begin{subfigure}[b]{0.36\textwidth}
        \centering
        \includegraphics[width=\textwidth]{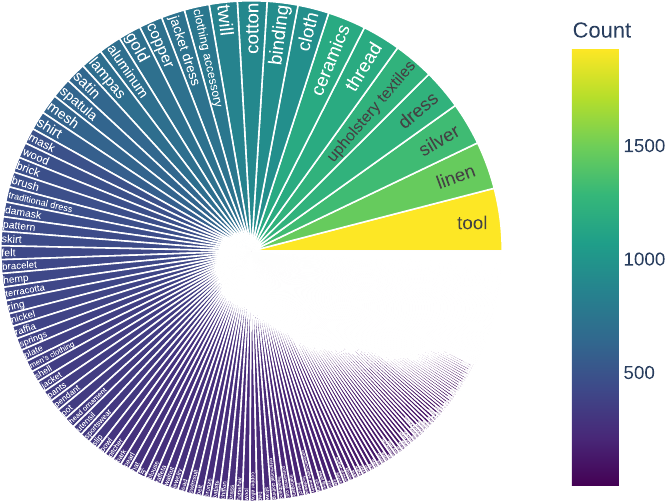}
        \caption{Validation set.}
        \label{fig:val_sunburst}
    \end{subfigure}
    \vfill
    \begin{subfigure}[b]{0.36\textwidth}
        \centering
        \includegraphics[width=\textwidth]{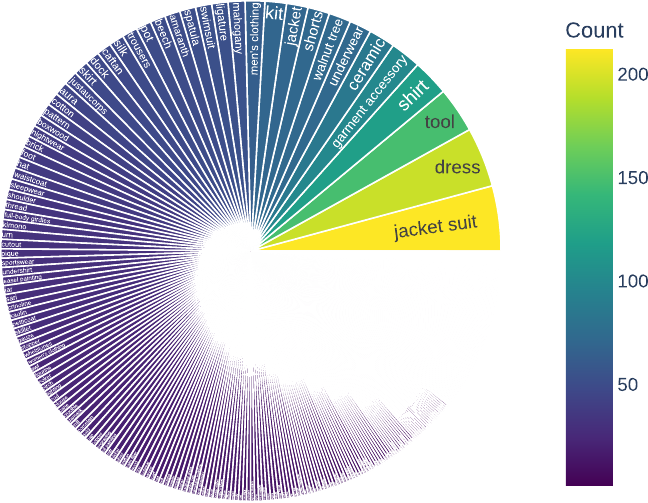}
        \caption{Inner test set.}
        \label{fig:test_id_sunburst}
    \end{subfigure}
    \hfill
    \begin{subfigure}[b]{0.36\textwidth}
        \centering
        \includegraphics[width=\textwidth]{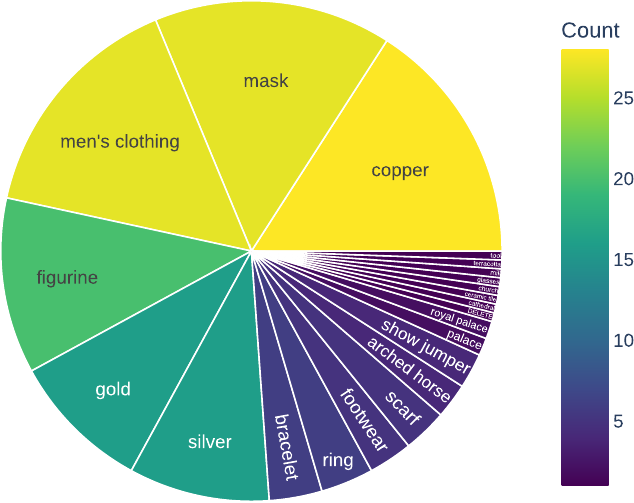}
        \caption{Outer test set.}
        \label{fig:test_ood_sunburst}
    \end{subfigure}
    \caption{Attribute labels' total frequency for the different partitions of the EUFCC-CIR dataset. Zoom-in for better visualization.}
    \label{fig:sunburst_plots_labels}
\end{figure}

\begin{figure}[h!]
    \centering
    \begin{subfigure}[b]{0.36\textwidth}
        \centering
        \includegraphics[width=\textwidth]{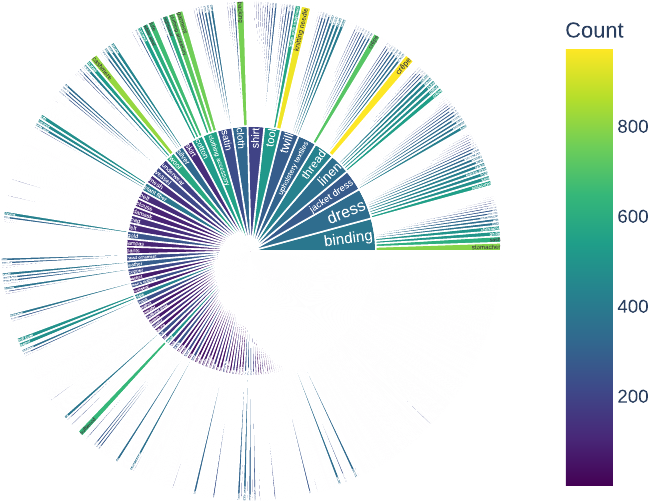}
        \caption{Training set.}
        \label{fig:train_sunburst_tuples}
    \end{subfigure}
    \hfill
    \begin{subfigure}[b]{0.36\textwidth}
        \centering
        \includegraphics[width=\textwidth]{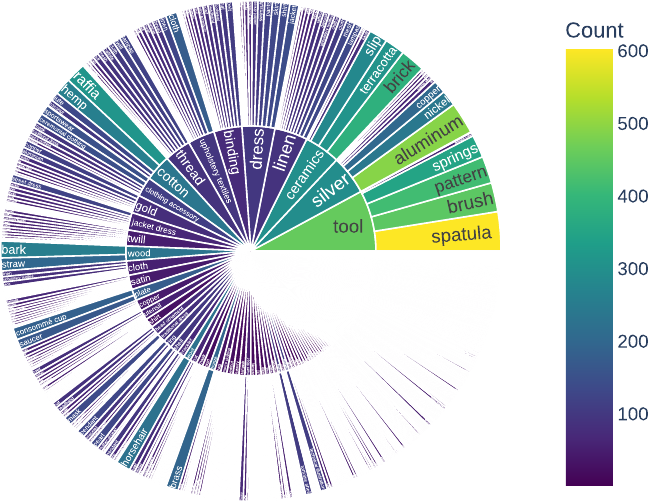}
        \caption{Validation set.}
        \label{fig:val_sunburst_tuples}
    \end{subfigure}
    \vfill
    \begin{subfigure}[b]{0.36\textwidth}
        \centering
        \includegraphics[width=\textwidth]{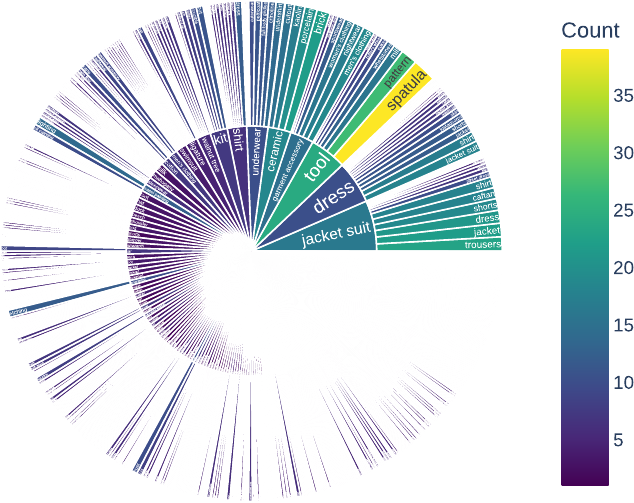}
        \caption{Inner test set.}
        \label{fig:test_id_sunburst_tuples}
    \end{subfigure}
    \hfill
    \begin{subfigure}[b]{0.36\textwidth}
        \centering
        \includegraphics[width=\textwidth]{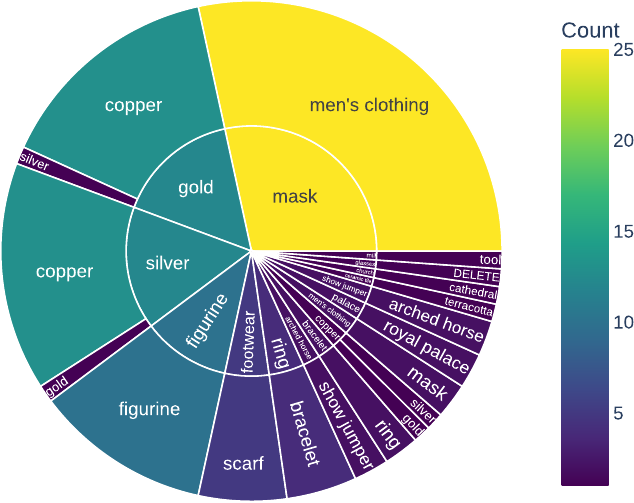}
        \caption{Outer test set.}
        \label{fig:test_ood_sunburst_tuples}
    \end{subfigure}
    \caption{Frequency of source-target attribute labels' pairs for the different partitions of the EUFCC-CIR dataset. The source\_element is represented in the inner circle and the target\_element in the outter circle. Zoom-in for better visualization.}
    \label{fig:sunburst_plots_pairs}
\end{figure}

\begin{figure}[htbp]
    \centering
    \begin{subfigure}[b]{0.45\textwidth}
        \centering
        \includegraphics[width=\textwidth]{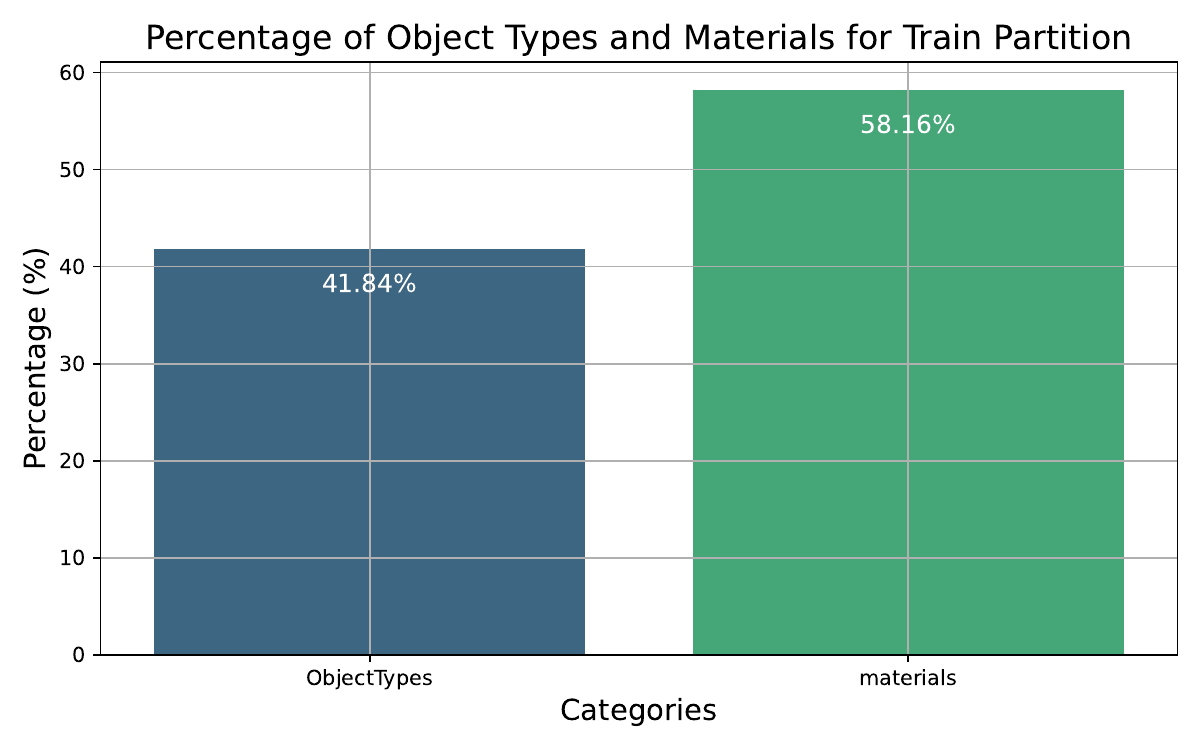}
        \caption{Training set.}
        \label{fig:train}
    \end{subfigure}
    \hfill
    \begin{subfigure}[b]{0.45\textwidth}
        \centering
        \includegraphics[width=\textwidth]{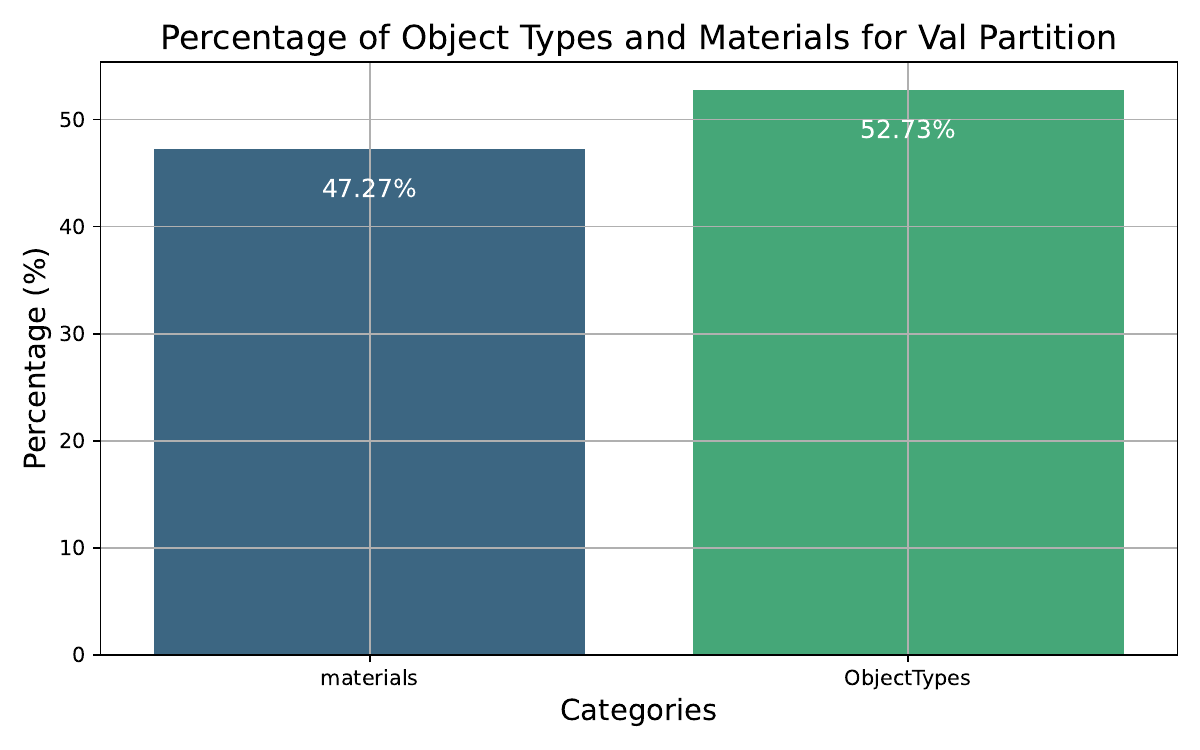}
        \caption{Validation set.}
        \label{fig:val}
    \end{subfigure}
    \vfill
    \begin{subfigure}[b]{0.45\textwidth}
        \centering
        \includegraphics[width=\textwidth]{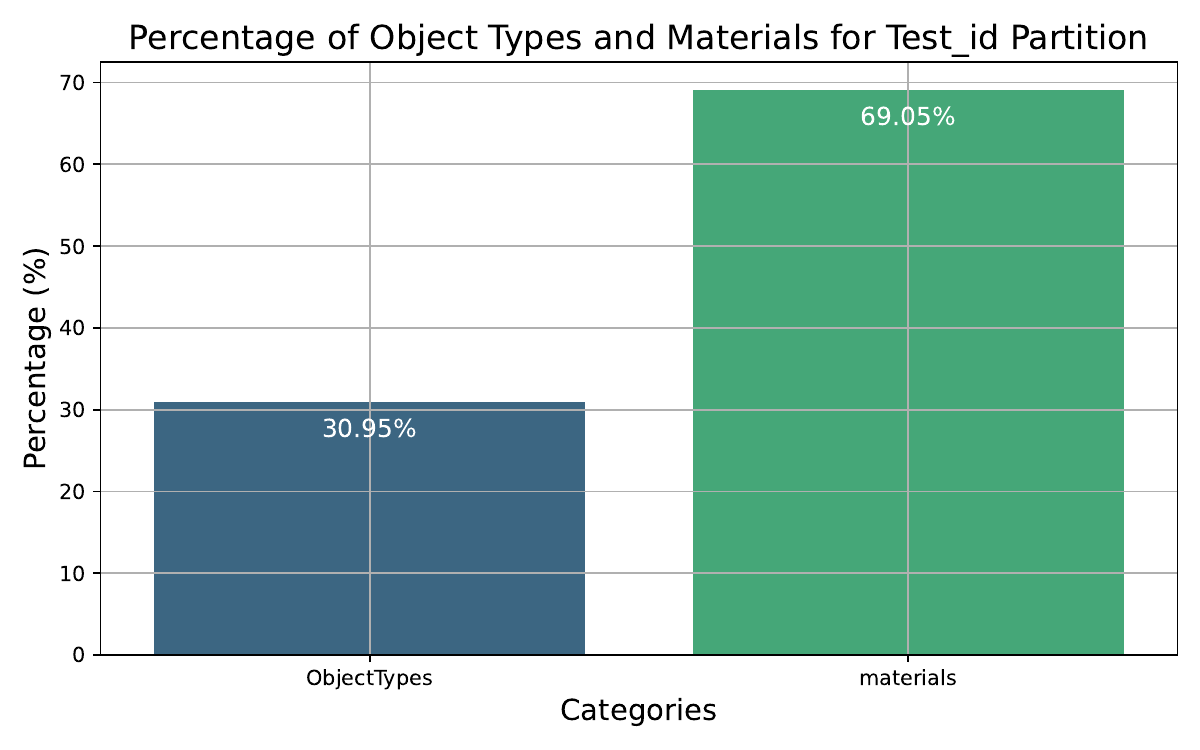}
        \caption{Inner test set.}
        \label{fig:test_id}
    \end{subfigure}
    \hfill
    \begin{subfigure}[b]{0.45\textwidth}
        \centering
        \includegraphics[width=\textwidth]{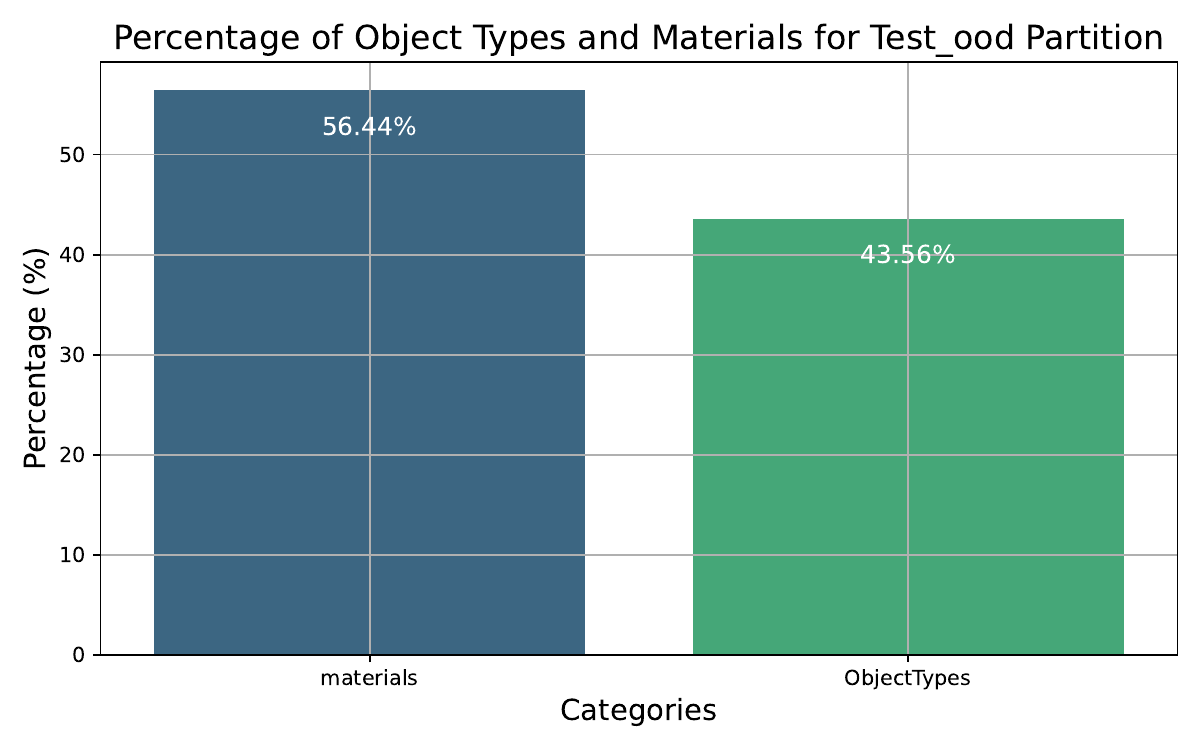}
        \caption{Outer test set.}
        \label{fig:test_ood}
    \end{subfigure}
    \caption{Percentage of triplets based on ``Object Types'' and ``Materials'' attributes for the different partitions of the EUFCC-CIR dataset.}
    \label{fig:partitions}
\end{figure}

The final dataset is composed of $2,648$ triplets for the test set (including both the inner and outer test sets), $149,686$ triplets for the training set, and $24,651$ triplets for the validation set. Figure \ref{fig:sunburst_plots_labels} shows the total frequency of attribute labels in each of the triplet subsets. Figure \ref{fig:sunburst_plots_pairs} shows the count frequency of source-target attribute label pairs, with the source element in the inner circle and the target element in the outer circle, for each of the triplet subsets. Finally, Figure \ref{fig:partitions} shows the percentages of triplets generated from different ``Object Types'' or different ``Materials'' in each set.

\section{Baseline Methods and Results}
\label{sec:baselines}

\begin{figure}[h]
    \centering
    \begin{subfigure}[b]{0.6\textwidth}
        \centering
        \includegraphics[width=\textwidth]{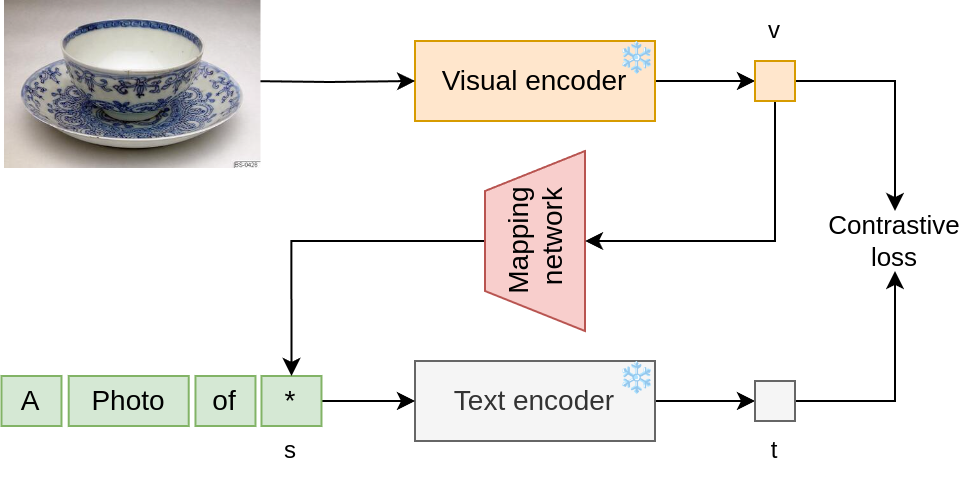}
        \caption{Training Process of the mapping network}
        \label{fig:inference}
    \end{subfigure}
    \vfill
    \begin{subfigure}[b]{\textwidth}
        \centering
        \includegraphics[width=\textwidth]{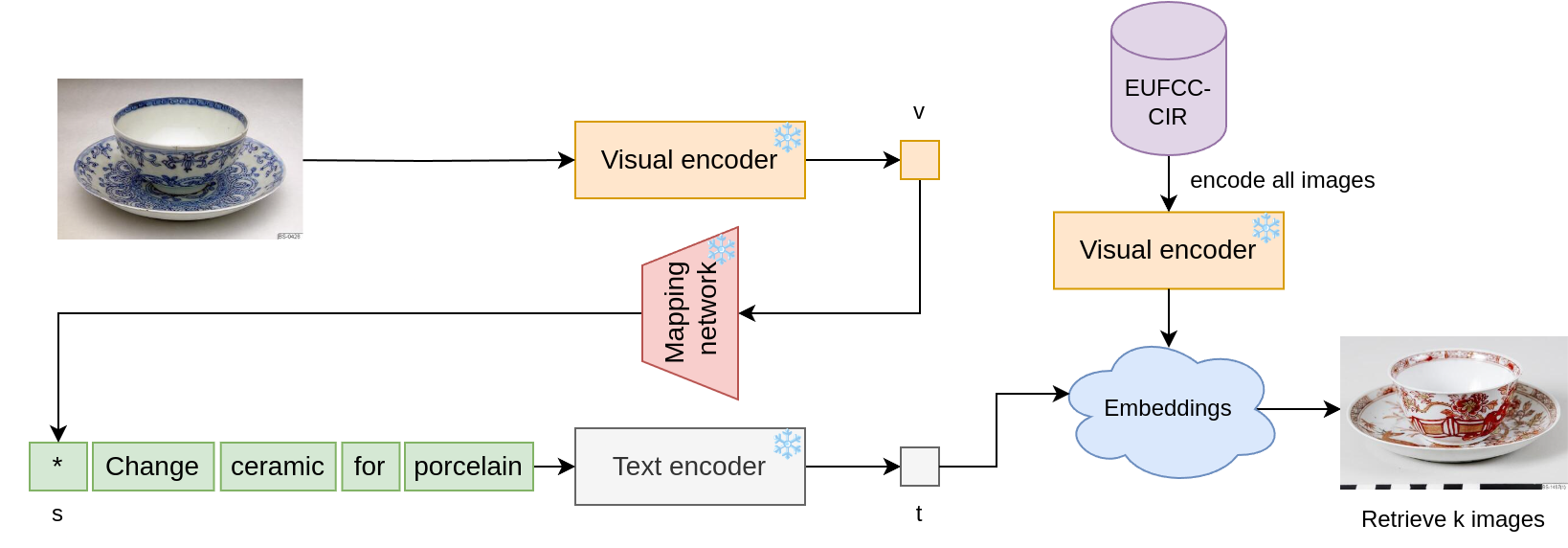}
        \caption{Inference Process to retrieve k images given an image and a query. The estimated token is used to fill in the given prompt sentence.}
        \label{fig:training}
    \end{subfigure}
    \caption{Overview of the  Pic2word~\cite{saito2023pic2word} method. (a) the mapping network is optimized to minimize the contrastive loss between the image embedding and the language embedding of a text template that includes the pseudo-word token $s$, creating a language embedding that is compatible with the embedding generated by the frozen CLIP language encoder. (b) once the mapping network is trained it can be used to project image queries into pseudo-word tokens so that the multi-modal CIR query is converted into a unimodal feature representation.}
    \label{fig:pic2word}
\end{figure}

In this section, we present the results of four different baseline methods based on zero-shot learning. These methods make use of different feature representations to address the composed image retrieval (CIR) task:

\begin{enumerate}[label=\Roman*., wide=0pt, itemsep=0em]
    \item \textbf{Visual Only}: We use the CLIP visual encoder to generate visual features from the query images, and then perform the retrieval task by calculating cosine similarities with the visual features of all images in the test set.
\vspace{1em}
    \item \textbf{Text Only}: We use the CLIP text encoder to generate query features from the paired textual descriptions only (ignoring the images), and then perform the retrieval task as before by calculating cosine similarities of the query text features with the visual features of all images in the test set.
\vspace{1em}
    \item \textbf{Mixture} (Averaged Features): This integrates both the visual and textual features into a single averaged mixture feature representation for each query.
\vspace{1em}
    \item \textbf{Pic2Word}~\cite{saito2023pic2word}: As explained in section~\ref{sec:sota} Pic2Word is a zero-shot compositional image retrieval method that combines pre-trained visual and text encoders from CLIP. As illustrated in Figure \ref{fig:pic2word} a mapping network is trained to convert the visual embedding of the input image into a corresponding pseudo-language token. This allows the multi-modal query to be transformed into a single modality (text). During training, the network is optimized to ensure the pseudo token faithfully represents the visual embedding. At test time, the predicted token is inserted into a template alongside the query text, and the resulting features are compared to candidate images.

\end{enumerate}

In our experiments we use three different text prompt templates: ``\texttt{* change {<}OLD{>} for {<}NEW{>}}'', ``\texttt{change * for {<}NEW{>}}'', and ``\texttt{* for {<}NEW{>}}''; where the symbol \texttt{*} represents the pseudo-language token, and the placeholders \texttt{{<}OLD{>}} and \texttt{{<}NEW{>}} respectively represent the attributes to be changed from the query image according to the EUFCC-CIR dataset ground-truth triplets (see section~\ref{sec:dataset}).

For all the baselines we use the \textit{ViT-B/16} model from the OpenCLIP repository~\cite{ilharco_gabriel_2021_5143773}. For the Pic2Word results, the mapping network is trained with the default parmeters from the original Pic2Word repository\footnote{\url{https://github.com/google-research/composed_image_retrieval}} using only the images of the EUFCC-340K training set (without the paired CIR text modifiers). Accordingly, all the provided results below are based on zero-shot learning as they are not trained using the EUFCC-CIR training set triplets. We expect supervised methods trained on the EUFCC-CIR training triplets to outperform these zero-shot baselines. 

\subsection{Results}

Table \ref{tab:combined_features} presents the outcomes of all previously described baselines on the EUFCC-CIR Inner and Outer test sets. The metrics used in this evaluation are Recall@1, Recall@5, Recall@10, and Recall@50, which represent the proportion of relevant instances retrieved within the top 1, 5, 10, and 50 results, respectively.

\begin{table}[h]
\centering
\caption{Performance comparison of different zero-shot baselines on the EFCC-CIR Inner and Outer test sets using various text prompt templates.}
\label{tab:combined_features}
\setlength{\tabcolsep}{2.25pt}
\begin{tabularx}{\textwidth}{lC|rrrr|rrrr}
\toprule
\textbf{Method} & \textbf{Text Prompt} & \multicolumn{4}{c}{\textbf{Inner test set}} & \multicolumn{4}{c}{\textbf{Outer test set}} \\ 
 & & \textbf{~R1} & \textbf{~R5} & \textbf{R10} & \textbf{R50} & \textbf{~R1} & \textbf{~R5} & \textbf{R10} & \textbf{R50} \\ \midrule
\multirow{1}{*}{Visual Only} & \textit{n.a.} & ~2.1 & ~8.2 & 13.5 & 30.3 & ~2.8 & ~4.4 & ~6.9 & 11.4 \\ \midrule
\multirow{3}{*}{Text Only} & \scriptsize\texttt{* change {<}OLD{>} for {<}NEW{>}} & 1.8 & 8.0 & 11.2 & 26.6 & 0.9 & 2.5 & 7.7 & 14.0 \\
 & \scriptsize\texttt{change * for {<}NEW{>}} & 0.7 & 5.5 & 8.2 & 23.9 & 0.3 & 1.3 & 5.3 & 8.9 \\
 & \scriptsize\texttt{* for {<}NEW{>}} & 0.1 & 5.5 & 9.2 & 23.8 & 0.3 & 5.3 & 5.6 & 8.4 \\ \midrule
\multirow{3}{*}{Mixture} & \scriptsize\texttt{* change {<}OLD{>} for {<}NEW{>}} & \textbf{3.5} & \textbf{10.2} & \textbf{15.7} & \textbf{34.2} & \textbf{3.0} & 4.6 & 9.8 & 14.3 \\
 & \scriptsize\texttt{change * for {<}NEW{>}} & 3.2 & 9.0 & 15.3 & 33.2 & 3.3 & \textbf{6.0} & \textbf{10.0} & 14.3 \\
 & \scriptsize\texttt{* for {<}NEW{>}} & 3.0 & 9.3 & 14.7 & 33.3 & 2.8 & 5.7 & 9.8 & 13.7 \\ \midrule
\multirow{3}{*}{Pic2Word~\cite{saito2023pic2word}} & \scriptsize\texttt{* change {<}OLD{>} for {<}NEW{>}} & 2.8 & 7.9 & 12.8 & 31.3 & 2.0 & 4.5 & 7.9 & \textbf{15.8} \\
 & \scriptsize\texttt{change * for {<}NEW{>}} & 2.8 & 8.0 & 14.0 & 32.5 & 3.1 & 5.5 & 8.9 & 15.4 \\
 & \scriptsize\texttt{* for {<}NEW{>}} & 2.7 & 8.0 & 13.2 & 31.3 & 3.1 & 5.6 & 8.9 & 14.2 \\ 

\bottomrule
\end{tabularx}
\end{table}

We appreciate that the mixture method (averaged features) outperforms all other baselines. This contradicts our initial assumption that the Pic2Word compositional features would yield the best performance. 
Across most ranks (R1, R5, R10, R50), the mixture method performed either slightly better or similarly to the Pic2Word method. The only exception being R50 on the Outer test set. These results indicate that a simple method like averaging features provides a reasonably robust zero-shot baseline for the EUFCC-CIR dataset.

\begin{figure}[t]
\vspace{-1em}
    \centering
    \includegraphics[width=\textwidth]{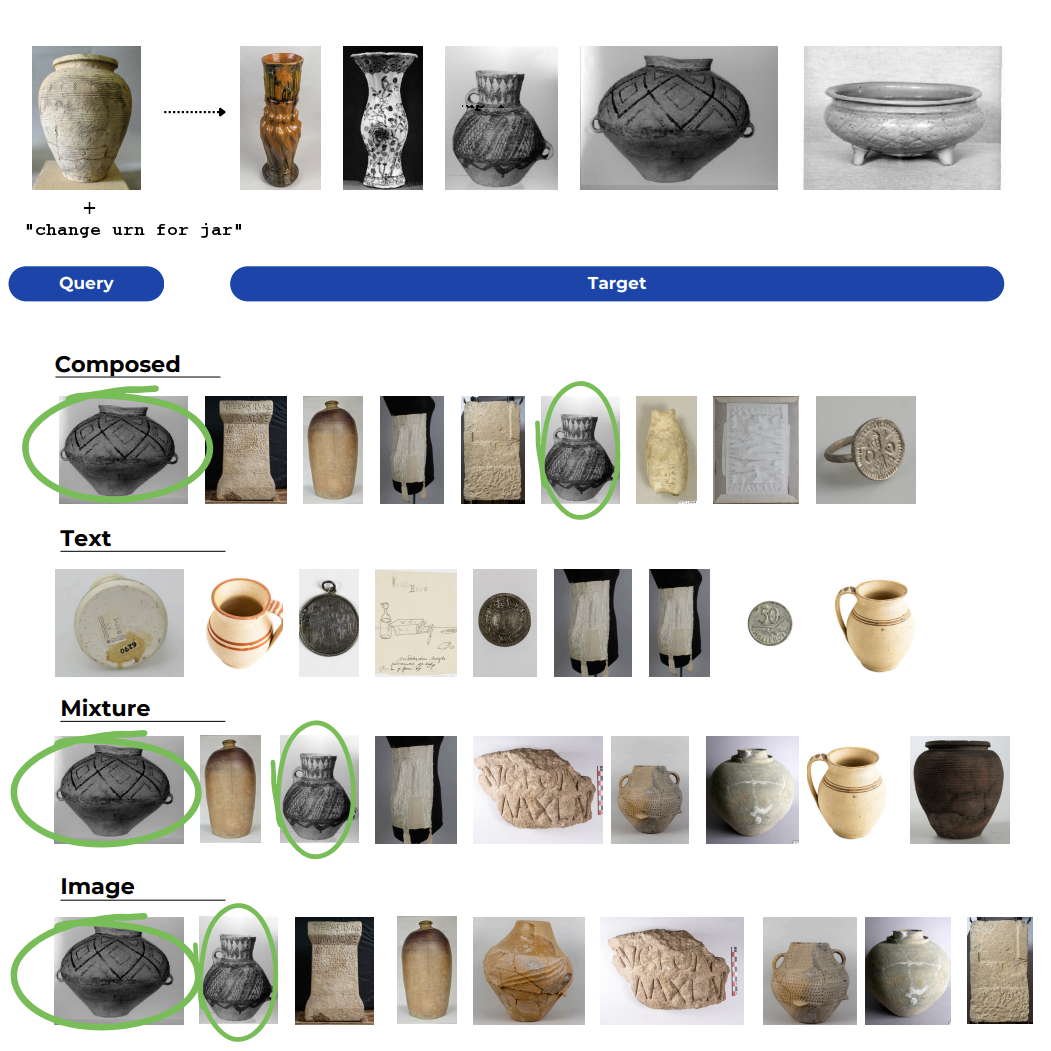}
    \caption{Zero-shot CIR qualitative results for the item \textit{ceramic\_$025196$} of the EUFCC-CIR test set for all baselines: text, vision, mixture, and composed (Pic2Word).}
    \label{fig:qualitatives2}
\end{figure}

\begin{figure}[h]
    \centering
    \includegraphics[width=\textwidth]{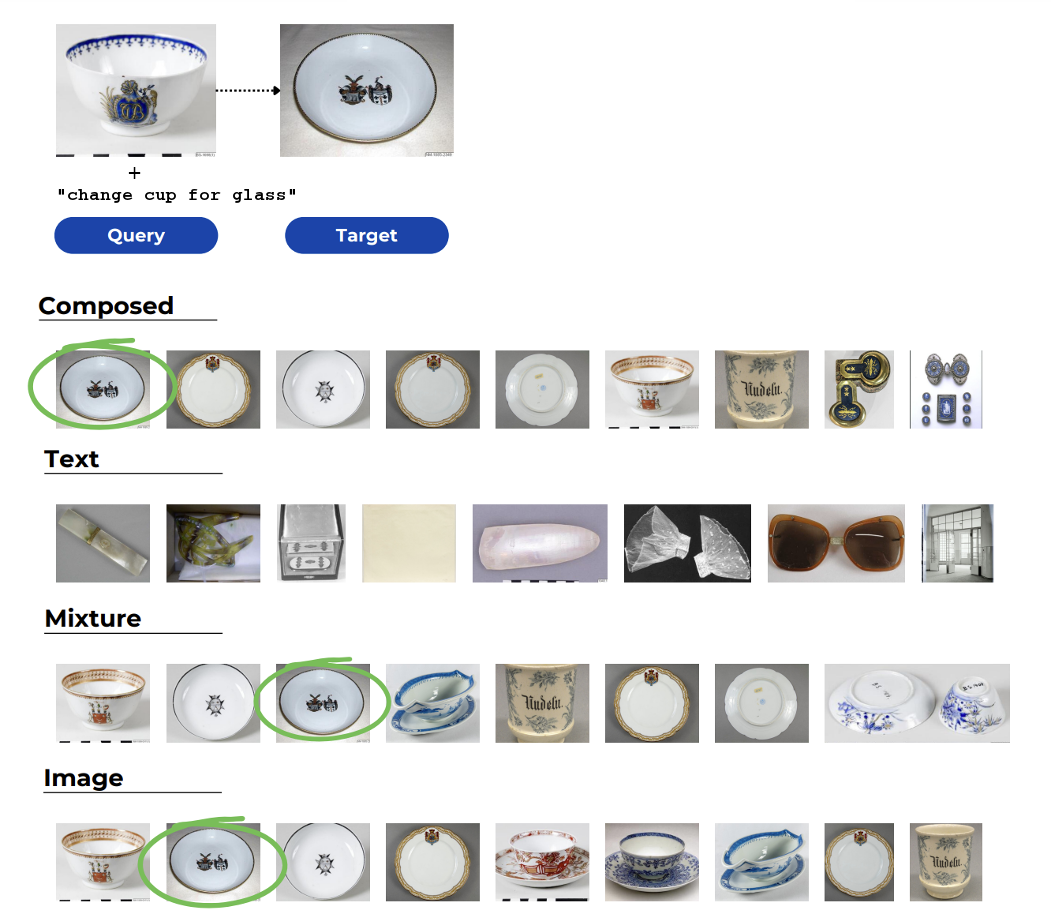}
    \caption{Zero-shot CIR qualitative results for the item \textit{ceramic\_$002498$} of the EUFCC-CIR test set for all baselines: text, vision, mixture, and composed (Pic2Word).}
    \label{fig:qualitatives1}
\end{figure}

Figures \ref{fig:qualitatives2} and  \ref{fig:qualitatives1} show qualitative retrieval results for two queries of the EUFCC-CIR test set using all baselines: text, vision, mixture, and composed (Pic2Word). In each example we show the query pair (image + text) and the ground-truth targets, as well as the top retrieved images with each method. We appreciate that text only features tend to provide non-relevant results, while visual only features have better alignment with the visual aspects of the queries, although sometimes do not incorporate the exact modifications from the query specification. Both Mixture and Compositional (Pic2Word) approaches were robust at balancing the visual and textual inputs in these examples.

\section{Conclusions}
\label{sec:conclusions}
In this paper a novel dataset has been presented, designed explicitly for Composed Image Retrieval (CIR) tasks within Galleries, Libraries, Archives, and Museums (GLAM) collections: EUFCC-CIR. Leveraging the large-scale EUFCC-340K dataset, an automated process and a set heuristic filters have been applied to create a rich and varied set of multimodal queries (image+text) and relevant images' set triplets, capturing meaningful variations of cultural artifacts.  A focus has been made on ``Object Type'' and ``Materials'' attributes, restricting the boundaries of the original EUFCC-340K subset splits while creating the triplets. 

On this new dataset we have evaluated several zero-shot baselines. The results highlight the potential of the EUFCC-CIR dataset for advancing research in composed image retrieval, especially for enriching user experience and enhancing research in the the digital humanities field. Future research might explore supervised learning (training from scratch or fine-tuning with the EUFCC-CIR training set), and extending the dataset to include more attributes or a more diverse array of cultural artifacts.

\section*{Acknowledgements}
This work has been supported by the ACCIO INNOTEC 2021 project Coeli-IA (ACE034/21/000084), and the CERCA Programme / Generalitat de Catalunya. Lluis Gomez is funded by the Ramon y Cajal research fellowship RYC2020-030777-I / AEI / 10.13039/501100011033.

%
%
\bibliographystyle{splncs04}
\bibliography{main}
\end{document}